\documentclass{article}

\usepackage{arxiv}

\usepackage[square,numbers]{natbib}
\usepackage[utf8]{inputenc} 
\usepackage[T1]{fontenc}    % use 8-bit T1 fonts
\usepackage{hyperref}       % hyperlinks
\usepackage{url}            % simple URL typesetting
\usepackage{booktabs}       % professional-quality tables
\usepackage{nicefrac}       % compact symbols for 1/2, etc.
\usepackage{microtype}     
\usepackage{amsfonts}       % blackboard math symbols
\usepackage{nicefrac}       % compact symbols for 1/2, etc.
\usepackage{microtype}  
\usepackage{amsmath}
\usepackage{amssymb}
\usepackage{amsthm}
\usepackage{graphicx}%add pictures

\usepackage[normalem]{ulem}
\useunder{\uline}{\ul}{}
\usepackage{titlesec}
\usepackage{lastpage}
\usepackage{xcolor}         % colors
\usepackage{bbm}
\usepackage{multirow}

\title{Do We Really Need to Learn Representations from In-domain Data for Outlier Detection?}

\author{
  Zhisheng Xiao \\
  Computational and Applied Mathematics\\
  University of Chicago\\
  Chicago, IL, 60637\\
  \texttt{zxiao@uchicago.edu} \\
   \And
  Qing Yan \\
  Department of Statistics\\
  University of Chicago\\
  Chicago, IL, 60637\\
  \texttt{yanq@uchicago.edu} \\
    \And
  Yali Amit \\
  Department of Statistics\\
  University of Chicago\\
  Chicago, IL, 60637\\
  \texttt{amit@marx.uchicago.edu} \\
}

\begin{document}

\maketitle

\begin{abstract}
  Unsupervised outlier detection, which predicts if a test sample is an outlier or not using only the information from unlabelled inlier data, is an important but challenging task. Recently, methods based on the two-stage framework achieve state-of-the-art performance on this task. The framework leverages self-supervised representation learning algorithms to train a feature extractor on inlier data, and applies a simple outlier detector in the feature space. In this paper, we explore the possibility of avoiding the high cost of training a distinct representation for each outlier detection task, and instead using a single pre-trained network as the universal feature extractor regardless of the source of in-domain data. In particular, we replace the task-specific feature extractor by one network pre-trained on ImageNet with a self-supervised loss. In experiments, we demonstrate competitive or better performance on a variety of outlier detection benchmarks compared with previous two-stage methods, suggesting that learning representations from in-domain data may be unnecessary for outlier detection.
\end{abstract}

\section{Introduction}
In order to make reliable decisions, deep learning models that are deployed for safety-critical applications, such as medical diagnosis \cite{mti2030047}, autonomous driving \cite{evtimov2017robust} and face authentication \cite{Wang2021DeepFR}, need to be able to identify whether the input data is significantly different from the training data. Such a task is called outlier detection or Out-of-distribution (OOD) detection. Previous approaches to outlier detection often rely on calibrating the classification confidence on labeled data \cite{hendrycks2016baseline,Liang2018EnhancingTR,Lee2018TrainingCC} or contrasting with known OOD samples \cite{bahat2018confidence, hendrycks2016baseline, ruff2019deep}. Unfortunately, the applicability of these methods is limited. In practice, machine learning models are used to tackle a much wider range of problems than classification, and therefore labeled data may not be available. In addition, 
the space of all possible OOD samples is typically huge, and hence using specific external data to represent OODs may introduce bias to the detector. Therefore, it is more promising to study \textit{unsupervised} outlier detection \cite{Bulusu2020AnomalousED}, where neither labels of in-domain data nor concrete examples of outliers are available. 

Detecting OOD samples with only unlabeled data from the training distribution is challenging. Previous approaches based on reconstruction \citep{pidhorskyi2018generative,zong2018deep} or density estimation \cite{Du2019ImplicitGA,Ren2019LikelihoodRF,xiao2020likelihood} do not obtain comparable performance with classifier based outlier detectors. More importantly, it is observed that density estimation with some probabilistic generative models may assign higher likelihoods to outliers than in-distribution
data \cite{nalisnick}, suggesting that there might be fundamental issues with this approach \cite{le2020perfect}. An alternative framework for unsupervised outlier detection can be summarized as a two-stage procedure, where in the first stage a neural network is used to extract a high-level representation of data and, in the second stage, an outlier detector is applied on the representation space. The key component of this framework is a good feature extractor that ensures the features of in-distribution data are clustered together, while keeping the features of outliers away from the cluster. Previous two-stage outlier detection methods require training the feature extractor on in-distribution data, either with a classification loss (if labels are available) \cite{ahuja2019probabilistic,lee2018simple} or a self-supervised loss \cite{sehwag2021ssd, tack2020csi, sohn2020learning}. 

In particular, the best feature extractors used in \cite{sehwag2021ssd, tack2020csi,sohn2020learning} are trained with a contrastive loss \cite{chen2020simple, he2020momentum,caron2020unsupervised}, which has led to tremendous progress in representation learning. However, training such representations can be difficult, as it may require large batch size \cite{chen2020simple}, long training epochs, and carefully designed data augmentation and training scheme \cite{chen2020simple,he2020momentum,chen2020big, tian2020makes,xiao2020should}. Note that in practice, outlier detection is often a \textit{side task}, where outliers are filtered out before entering the networks designed for the main task. Therefore, the high cost of training feature extractors for outlier detection is undesirable. 

Considering the fact that pre-trained feature extractors on some reference datasets (such as ImageNet in the image domain) are easily accessible,\footnote{For example, pre-trained weights obtained from supervised training and a variety of self-supervised training algorithms on ImageNet are publicly available at \url{https://github.com/facebookresearch/vissl/blob/master/MODEL_ZOO.md}} and they are shown to be effective in transferring knowledge to other datasets \cite{marcelino2018transfer,grill2020koray,lu2021contrastive,azizi2021big}, we ask a natural but, to our knowledge, unexplored question: can we directly use feature extractors trained on a reference dataset without any further training to detect outliers for any source of in-distribution data? In this paper, we give an affirmative answer to this question. Specifically, we focus on outlier detection in the image domain, and show that self-supervised feature extractors pre-trained on ImageNet lead to comparable or even better performances than state-of-the-art two-stage outlier detectors, which require expensive training to obtain representations of in-distribution data. In other words, we obtain an universal outlier detector without training on particular in-distribution data. This means that, given the public availability of pre-trained feature extractors, we can detect outlier samples \textit{for free}. 

\section{Backgrounds and Our Approach} \label{bg and method}
In this section, we first briefly introduce the problem formulation, and provide the necessary background on self-supervised representation learning, since it is an important component of our approach and related methods. Then we provide a detailed review of the two-stage outlier detection framework, and position our approach in that framework.

\textbf{Problem Formulation. }Suppose we have a set of $N$ unlabeled training samples $\left\{\boldsymbol{\mathbf{x}}_{i}\right\}_{i=1}^{N}$ drawn from some underlying data distribution $\mathbf{x}_i \sim p(\mathbf{x})$. The goal of outlier detection is to decide whether a test sample $\mathbf{x}$ is an outlier, which means that $\mathbf{x}$ has low density under true data distribution $p(\mathbf{x})$. Since $p(\mathbf{x})$ cannot be easily modeled, usually we define a certain score
function $s(\mathbf{x})$ where a higher (or lower) value indicates that $\mathbf{x}$ is more likely from in-distribution. There are two common outlier detection tasks: \textbf{OOD detection}, where outliers are collected from different sources from in-distribution data, with different semantics and texture (for example, CIFAR-10 vs. SVHN), and \textbf{anomaly detection}, where outliers have similar texture but different semantics \cite{pang2020deep}. Typically anomaly detection is evaluated on a multi-class dataset, where the goal is to detect samples from other classes given one particular class as in-distribution. While many proposed methods are particularly designed for one task \cite{perera2019deep, wang2019effective,golan2018deep}, we will show that our method is effective in both OOD detection and anomaly detection.

\textbf{Background: Self-supervised Representation Learning. }Earlier self-supervised methods rely on using auxiliary handcrafted prediction tasks to learn representations \cite{doersch2017multi,larsson2016learning,noroozi2016unsupervised, gidaris2018unsupervised}. These methods have been largely outperformed by contrastive methods, which learn representations by discriminating between individual instances:
given an augmented view of an image (such as translation, scaling, etc.), the network learns to discriminate between another augmented view of the same image, and augmented views of different images, commonly called negative samples. Formally, for each image in a batch of size $N$, a pair of augmented views are created, resulting in $2N$ data points. For each pair $(\mathbf{x}_i, \mathbf{x}_j)$ of positive images, the remaining $2(N-1)$ augmented images serve 
as negative examples. Each augmented image $\mathbf{x}$ is mapped through the feature extractor $f$ and projection head $h$, resulting in $\mathbf{z} = h(f(\mathbf{x}))$, and the loss for the batch can be written as 
\begin{align}\label{simclr_loss}
    \mathcal{L} =\frac{1}{2 N} \sum_{i=1}^{2 N}-\log \frac{e^{\text{sim}(\mathbf{z}_i, \mathbf{z}_j) / \tau}}{\sum_{k=1}^{2 N} \mathbbm{1}(k \neq i) e^{\text{sim}(\mathbf{z}_i, \mathbf{z}_k)/ \tau}},
\end{align}
where $\text{sim}(\cdot)$ is a measurement of similarity, such as the cosine similarity $\text{sim}(\mathbf{u}, \mathbf{v}) = \mathbf{u}^{\top} \mathbf{v} /\|\mathbf{u}\|\|\mathbf{v}\|$ \cite{chen2020simple}, and $\tau$ is the temperature. It is shown that the representations obtained from contrastive learning yield good performance in linear classification and a variety of downstream tasks \cite{kotar2021contrasting}. Besides the simplest contrastive algorithm presented above, several related algorithms have also proven to be effective in learning representations \cite{he2020momentum,chen2020improved,grill2020koray,caron2020unsupervised,zbontar2021barlow}.

\subsection{Two-stage Outlier Detection} \label{two stage review}
\textbf{First Stage: }In the first stage, in-distribution samples $\left\{\mathbf{x}_{i}\right\}_{i=1}^{N}$ are mapped to the representation space by a feature extractor network $f$, and the resulting representations $\mathbf{z}_i = f(\mathbf{x}_i)$ are collected. When there is no label available, previous methods train self-supervised feature extractors on in-distribution data \cite{sehwag2021ssd,sohn2020learning,tack2020csi}. In particular, it is observed in their experiments that feature extractors trained by contrastive loss lead to better results than those trained by handcrafted self-supervised tasks. 

Training self-supervised representations can be very expensive. For example, previous methods typically need 1000 or even more than 2000 training epochs on in-distribution data \cite{tack2020csi,sohn2020learning}. Moreover, they introduce modifications to the contrastive loss for better outliter detection performance. For example, \cite{tack2020csi} made a major modification to the representation learning scheme in equation \eqref{simclr_loss}: instead of simply considering augmented samples as positive to each other, it designs a specific set of transformations, called \textit{shifting transformations}, and samples augmented with transformations in this set are treated as negative samples. A similar idea is  used in \cite{sohn2020learning}. The intuition is that data applied with certain hard transformations can be treated as representative of OOD samples, thus contrasting against them would lead to improved OOD detection. However, such a modification requires careful selection of shifting transformations, and the resulting representation is less useful for other downstream tasks other than OOD detection. The high training cost and the subtlety of modifying the loss together motivate us to use pre-trained representations as a universal feature extractor.

\textbf{Second Stage: }
In the second stage, after collecting the set of features $\left\{\mathbf{z}_{i}\right\}_{i=1}^{N}$, a test sample $\mathbf{x}_{\text{test}}$ is mapped to the feature space using $f$ to obtain $\mathbf{z}_{\text{test}}$, and a simple outlier detector is used to compute the detection score $s(\mathbf{x}_{\text{test}})$ in the feature space by treating $\left\{\mathbf{z}_{i}\right\}_{i=1}^{N}$ as in-distribution data. The motivation is that it is much easier to quantify distance between features, which are relatively low-dimensional vectors, than original data with high dimensionality and complicated structure.

A variety of simple outlier detectors have been used in the second stage. They can be roughly classified into non-parametric and parametric detectors. Nonparametric detectors do not assume a parametric density on $\left\{\mathbf{z}_{i}\right\}_{i=1}^{N}$. For example, \cite{sohn2020learning} propose to use One-class SVM \cite{scholkopf1999support} and Kernel Density Estimation (KDE) to obtain the detection score. \cite{tack2020csi} defines the detection score by the largest cosine similarity between $\mathbf{z}_{\text{test}}$ and the training feature set. Conversely, parametric detectors develop a parametric generative model on $\left\{\mathbf{z}_{i}\right\}_{i=1}^{N}$. Perhaps the simplest parametric detector models $\left\{\mathbf{z}_{i}\right\}_{i=1}^{N}$ by a single multivariate Gaussian distribution, and uses the Mahalanobis distance as the detection score. \cite{sehwag2021ssd} further develops a cluster-conditioned detection method in the feature space, where it first partitions the features for training data in $M$ clusters, and then it models features in each cluster independently as multivariate Gaussian, and the score is the minimum Mahalanobis distance to the center of clusters:
\begin{align} \label{maha}
    s(\mathbf{x}_{\text{test}}) =\min _{m}\left(\mathbf{z}_{\text{test}} -\mathbf{\mu}_{m}\right)^{T} \mathbf{\Sigma}_{m}^{-1}\left(\mathbf{z}_{\text{test}}-\mathbf{\mu}_{m}\right), \quad \mathbf{z}_{\text{test}} = f(\mathbf{x}_{\text{test}})
\end{align}
where $\mu_{m}$ and $\Sigma_{m}$ are the sample mean and sample covariance of the $m^{th}$ cluster of the features of in-distribution data. In \cite{lee2018simple}, based on the shared covariance assumption of linear discriminative analysis, $\Sigma_{m}$ is replaced by a shared covariance matrix $\Sigma$.

\subsection{Outlier Detection with Pre-trained Representation} \label{components}
Now we introduce the details of our approach and position it in the two-stage outlier detection framework.

\textbf{Representation Learning Algorithm.} We replace the feature extractors trained on different in-distribution data in previous two-stage methods by a single feature extractor pre-trained on ImageNet. Some previous work such as \cite{sohn2020learning} include extracting features by a classifier trained on ImageNet as a baseline, and they show that ImageNet classifiers lead to worse results than their proposed representation learning methods, highlighting the importance of learning representations from in-domain distributions. However, we believe the main reason is that representations obtained from classifiers mainly keep label-related information on ImageNet, which may not be useful for detecting outliers for other sources of data. In contrast, self-supervised representation learning develops a richer understanding of semantics, and absence of such semantics in outlier samples can cause them to lie far away in the feature space. Therefore, we propose to use pre-trained self-supervised representations on ImageNet for outlier detection.  Since the representations are publicly available, we can try different self-supervised training algorithms with no additional cost. We compare several state-of-the-art self-supervised learning algorithms, including SimCLR \cite{chen2020simple,chen2020big}, MoCo \cite{he2020momentum,chen2020improved}, SwAV \cite{caron2020unsupervised}, and BYOL \cite{grill2020koray}. 

\textbf{Network Structure. }While there are ImageNet pre-trained classifiers with different network structures, most of the available representations learned by self-supervised training come with a ResNet-50 (or its wider variants) \cite{he2016deep} backbone. For lower computational and memory cost during testing, we mainly use the original ResNet-50 structure.
    
\textbf{Outlier detector in the feature space. }We choose to build a parametric OOD detector in the feature space. Parametric models have huge advantage in computation during testing, as non-parametric models such as KDE and nearest neighbors require pair-wise computation with each element in the training set. 
Following \cite{sehwag2021ssd,lee2018simple}, we first partition the training features into $M$ clusters by fitting a Gaussian mixture model, and the minimum  Mahalanobis distance defined in equation \eqref{maha} is used to detect outliers. We empirically find a tied covariance matrix as in \cite{lee2018simple} leads to better performances.
    
\textbf{Data Pre-processing. }Since we use representations trained on ImageNet for outlier detection tasks with different sources, we need to pre-process the input data to fit the required size. We will compare several simple resizing methods, such as nearest neighbor, bi-linear and cubic interpolation. In addition, we explore adding small perturbations to the input similar to \cite{Liang2018EnhancingTR,lee2018simple}. Specifically, we pre-process a test input $\mathbf{x}$ to obtain $\widehat{\mathbf{x}}$ by adding a small controlled noise:
    \begin{align}\label{perturbation}
        \widehat{\mathbf{x}}=\mathbf{x}+\varepsilon \operatorname{sign}\left(\nabla_{\mathbf{x}} s(\mathbf{x})\right)=\mathbf{x}-\varepsilon \operatorname{sign}\left(\nabla_{\mathbf{x}}\left(f(\mathbf{x})-\mu_{\widehat{m}}\right)^{\top} \mathbf{\Sigma}^{-1}\left(f(\mathbf{x})-\mu_{\widehat{m}}\right)\right),
    \end{align}
    where $\widehat{m}$ is the index of the closest cluster, $\mathbf{\Sigma}$ is either the tied covariance matrix or the covariance for cluster $\widehat{m}$, and $\epsilon$ is the magnitude of perturbation. Intuitively, such a perturbation will make in- and out-of-distribution samples more separable \cite{lee2018simple}.

\section{Related Work}
\label{RW}
\textbf{Unsupervised Outlier Detection: }Most previous approaches can be categorized as based on either density, reconstruction or feature distance. Density-based methods use the likelihood of test samples, usually obtained from a generative model trained on training data, as a decision score. However, recent studies show that on complex data such as images, many deep generative models will assign high likelihood to certain OOD samples \cite{nalisnick, choi2018waic,kirichenko2020normalizing}. Several works proposed to use alternative likelihood-based scores \cite{Ren2019LikelihoodRF,nalisnick2019detecting, serra2019input,choi2018waic, xiao2020likelihood, havtorn2021hierarchical} and show improved performance.
Reconstruction-based methods train an encoder-decoder structure to reconstruct the training data, and they use the reconstruction loss as a decision score \cite{zong2018deep, pidhorskyi2018generative, denouden2018improving, perera2019deep}. The main idea is that the network would obtain a worse reconstruction for OOD samples, which is shown to be false in certain cases \cite{nalisnick,denouden2018improving}. Note that both density- or reconstruction-based methods are largely outperformed by classifier based outlier detectors \cite{Liang2018EnhancingTR, Lakshminarayanan2017SimpleAS, Lee2018TrainingCC}. Feature distance based methods extract the neural representation of training data, and detect OOD samples based on the distance in the representation space \cite{ sehwag2021ssd,tack2020csi,sohn2020learning}. We have reviewed this framework in Section \ref{two stage review}. These methods obtain the best results in unsupervised outlier detection, and greatly reduce the performance gap with supervised outlier detection. 

\textbf{Other Outlier Detection Methods with Self-supervised Learning: }Some other outlier detection methods have a self-supervised learning component, but they are not based on feature distance. For example, \cite{hendrycks2019using,winkens2020contrastive,liu2020hybrid} use self-supervised loss in conjunction with supervised cross-entropy loss
to improve OOD detection. \cite{bergman2020classification,golan2018deep} train a network to predict certain geometric transformations, and detect outliers by the prediction accuracy. 

\textbf{Feature Distance with ImageNet Pre-trained Networks:  }An important motivation of our work is the success of using ImageNet pre-trained features to define a proper distance for images from other sources. For instance, features obtained from ImageNet classifiers are used to define scores such as IS and FID \cite{heusel2017gans,salimans2016improved,sajjadi2018assessing} that measure the distance between sets of arbitrary images, and they have been widely used for assessing the sample quality of generative models. Recently, \cite{morozov2021on} propose to compute the FID score based on self-supervised representations, and show that the resulting score better aligns with human perceptual quality.\footnote{Strictly speaking, the resulting score should not be called FID, as the network is no longer the InceptionV3 \cite{szegedy2016rethinking}, but we still use the name for convenience.} ImageNet pre-trained classifiers are also shown to be good at measuring the distance between a pair of images \cite{zhang2018context,johnson2016perceptual}, and hence can be used to define an effective reconstruction loss \cite{esser2020taming}.

\section{Results}
In this section, we evaluate our proposed method on a variety of outlier detection tasks.
In particular, we study OOD detection, where the in-distribution samples come from a certain dataset and outliers come from other datasets, and anomaly detection, where in a multi-class dataset, images from one class are given as inlier and those from remaining classes are given as outlier. Unlike many previous works which only consider outlier detection on low-resolution images (typically $32 \times 32$), we will evaluated our method on both small and large images. For all experiments (except the ablation study that compares different representation learning algorithms), we use the ResNet-50 trained by \textbf{SimCLRv2} \cite{chen2020big} on ImageNet as the feature extractor. Experimental settings are presented in Appendix \ref{exp detail}. Details on downloading the pre-trained model can be found in Appendix \ref{download}. 

\textbf{Evaluation. }Similar to almost all other outlier detectors, our method outputs a decision score. Once such a scoring function has been learned, a classifier can be constructed by specifying a threshold. A useful performance metric is the area under the Receiver Operating Characteristic (AUROC$\uparrow$), which measures the quality of the trade-off of different thresholds \cite{roc, roc2}. Throughout the paper, we mainly use AUROC as the evaluation metric. %while area under the Precision Recall Curve (AUPRC$\uparrow$) and the False Positive Rate at 95\% True Positive Rate (FPR95$\downarrow$) will be used in addition.

\subsection{OOD Detection on Unlabeled Multi-class Datasets}
We largely follow \cite{tack2020csi} to choose the datasets for benchmarking our method on the OOD detection task. For low-resolution images, we mainly use CIFAR-10 \cite{krizhevsky2009learning} as in-distribution data, and consider the following datasets as out-of-distribution: SVHN \cite{netzer2011reading}, resized LSUN and ImageNet \cite{Liang2018EnhancingTR} (and their fixed version, see Appendix I of \cite{tack2020csi}), and CIFAR-
100 \cite{krizhevsky2009learning}. For high-resolution images, we use ImageNte-30 \cite{hendrycks2019using}, CUB-200 \cite{wah2011caltech}, Dogs \cite{khosla2011novel}, Pets \cite{parkhi2012cats}, Flowers \cite{nilsback2006visual}, Places \cite{zhou2017places}, and Caltech-256 \cite{griffin2007caltech}, among which different inlier and outlier pairs are considered. 

In Table \ref{cifar ood}, we present results of OOD detection with CIFAR-10 as in-distribution data. We compare our methods against previous unsupervised OOD detectors, either based on deep generative models or self-supervised feature extractors trained on in-distribution data. We observe that our method significantly outperforms prior generative model based methods, and is competitive with SOTA self-supervised methods. In addition, from the last two lines, we observe that  the feature extractor trained with self-supervised loss on ImageNet performs significantly better than the feature extractor trained with classification loss on ImageNet.

Next, we study OOD detection on higher resolution images, with different inlier-outlier pairs. Note that previous works only consider ImageNet-30 as the inlier, and we will compare with them in Appendix \ref{imagenet-30 benchmark}. In Table \ref{large image ood}, we compare the results of our proposed method with the baseline where the feature extractor is pre-trained by the classification task on ImageNet. We find that our method is effective in all inlier-outlier pairs, and obtain nearly perfect results in many instances. Similar to the observation in CIFAR-10 experiments, the pre-trained self-supervised representation largely outperforms the pre-trained classifiers, especially on difficult tasks such as Dogs vs. Pets and Caltech-256 vs Places. 

\begin{table}
  \caption{AUROC ($\%$) of various unsupervised OOD detection methods with CIFAR-10 as in-distribution. LSUN(F) and ImageNet(F) correspond to the fixed version of LSUN and ImageNet introduced in \cite{tack2020csi}, where they fixed the resize issue of resized LSUN and ImageNet datasets produced by broken image resize operations that contain artificial noise.}
  \label{cifar ood}
  \centering \small
  \resizebox{\textwidth}{!}{
 \begin{tabular}{lllllllll}
 \toprule
                                          &                           & SVHN  & LSUN  & ImageNet & LSUN (F) & ImageNet (F) &CIFAR-100 \\
\midrule
\multirow{6}{2cm}{\textbf{Generative Models}}        & Glow                      & 8.3   & -     & 66.3    &-& -& 58.2         \\
                                          & EBM \cite{Du2019ImplicitGA}                       & 63.0  &    -   &    -    &-&-  & 50.0         \\
                                          & VAEBM \cite{xiao2021vaebm}                    & 83.0  &   -    &  -     &-&-   & 62.0         \\
                                          & Input Complexity \cite{serra2019input}         & 95.0  &   -    & 71.6   &-&-  & 73.6             \\
                                          & Likelihood Ratio  \cite{Ren2019LikelihoodRF}        & 91.2  &   -    &  -        &-&-&     -            \\
                                          & Likelihood Regret \cite{xiao2020likelihood}        & 87.5 & 69.1 &   -      &-&- &     -      \\
\midrule
\multirow{4}{2cm}{\textbf{Self-supervised Training}} & Rot + Trans  \cite{hendrycks2019using}             & 97.8  & 89.2  & 90.5    &81.6&86.7 & 79.0       \\
                                          & GOAD \cite{bergman2020classification}                     & 96.3  & 89.3  & 91.8   &78.8&83.3  & 77.2       \\
                                          & CSI \cite{tack2020csi}                     & \textbf{99.8}  & 97.5  & 97.6   &90.3&\textbf{93.3}  & 89.2         \\
                                          & SSD \cite{sehwag2021ssd}                       & 99.6  &   -   &    -    &-&-  & \textbf{90.6}         \\
\midrule
\multirow{2}{2cm}{\textbf{ImageNet Pre-trained}}              & Supervised    & 86.3  & 76.9  & 80.3    &69.9&76.8 & 70.4             \\
                                          & \textbf{Self-supervised (ours)} & 98.3  & \textbf{98.5}  & \textbf{98.6}    &\textbf{92.9}& 91.8& 81.7      \\
\bottomrule
\end{tabular}}
\end{table}

\begin{table}

  \caption{AUROC ($\%$) on various high resolution image datasets. Each row contains an in-lier dataset while each column contains an outlier dataset. For each grid in the table, \textbf{Top: }using features extracted by a ResNet-50 pre-trained with SimCLRv2, and \textbf{Bottom: }using features extracted by a ResNet-50 pre-trained with classification task.}
  \label{large image ood}
  \centering
 \begin{tabular}{lllllllll}
 \toprule
                              & ImageNet-30                  & CUB                          & Dogs                         & Flowers                                               & Pets                         & Places                       & Caltech                   \\ \midrule
\multirow{2}{1.5cm}{ImageNet-30} & \multirow{2}{1cm}{-}           & \multirow{2}{1cm}{99.8 74.5} & \multirow{2}{1cm}{99.5 95.4} & \multirow{2}{1cm}{98.1 85.1} & \multirow{2}{1cm}{99.7 92.5} & \multirow{2}{1cm}{80.0 75.4} & \multirow{2}{1cm}{89.2 80.9} \\
                             &                              &                              &                              &                              &                              &                              &                              &                              \\
                             \midrule
\multirow{2}{1.5cm}{Dogs}        & \multirow{2}{1cm}{99.9 95.5} & \multirow{2}{1cm}{99.9 96.5} & \multirow{2}{1cm}{-}           & \multirow{2}{1cm}{99.9 98.7} &  \multirow{2}{1cm}{94.5 72.4} & \multirow{2}{1cm}{99.7 94.4} & \multirow{2}{1cm}{99.2 95.2} \\
                             &                              &                              &                              &                              &                              &                              &                              &                              \\
                             \midrule
\multirow{2}{1.5cm}{Places}  & \multirow{2}{1cm}{99.4 89.4} & \multirow{2}{1cm}{99.8 87.8} & \multirow{2}{1cm}{99.0 95.7} & \multirow{2}{1cm}{97.4 86.4} &  \multirow{2}{1cm}{99.8 93.6} & \multirow{2}{1cm}{-}           & \multirow{2}{1cm}{93.4 84.1} \\
                             &                              &                              &                              &                              &                              &                              &                              &                              \\
                             \midrule
\multirow{2}{1.5cm}{Caltech} & \multirow{2}{1cm}{95.1 83.5} & \multirow{2}{1cm}{98.9 63.8} & \multirow{2}{1cm}{96.0 89.6} & \multirow{2}{1cm}{85.6 60.7} &  \multirow{2}{1cm}{99.2 91.5} & \multirow{2}{1cm}{74.3 61.9} & \multirow{2}{1cm}{-}           \\
                             &                              &                              &                              &                              &                              &                              &                              &                             \\
                             \bottomrule
\end{tabular}
\end{table}

\subsection{Anomaly Detection on Unlabeled One-class Datasets}
For anomaly detection tasks, we follow \cite{golan2018deep} to choose four image datasets in our experiments: CIFAR-10 (contains 10 classes) and CIFAR-100 (contains 100 classes, but we follow the common setting to group them into 20 super-classes) \cite{krizhevsky2009learning}, Fashion-MNIST (contains 10 classes) \cite{xiao2017fashion} and Cats-vs-Dogs (contains 2 classes) \cite{elson2007asirra}. We employ a one-vs-all evaluation scheme in each experiment. Consider a dataset with $C$ classes, from which we create $C$ different experiments, one for each class $c$. For each experiment, we collect the features of training examples belong to class $c$, and compute the statistics of the set of features. During testing, we use the images of class $c$ in the test set as inliers, and randomly sample equal number of images from the other $C-1$ classes in the test set as outliers. The resulting AUROCs for the $C$ experiments are averaged as the final metric. 

We report the above final metric in Table \ref{one class all table}, and compare our method with various methods based on self-supervised learning. Note that one-class anomaly detection is a more difficult task than OOD detection, as we have less training data and the outliers share similar texture with the inliers. Nevertheless, our proposed method is highly effective, achieving the best performances on two out of four tasks, while being slightly behind the best methods on the other two tasks. In particular, our method largely outperforms previous ones on Cat-vs-Dog, which is the only one-class anomaly detection task for high-resolution images.

\begin{table}
  \caption{Mean AUROC ($\%$) for one-class classification AUCs averaged over outlier classes and over 5 runs. We omit the standard deviations as they are small.}
  \label{one class all table}
  \centering
 \begin{tabular}{lllllll}
 \toprule
                                          &                           & CIFAR-10  & CIFAR-100  & f-MNIST & Cat-vs-Dog \\
\midrule

\multirow{9}{2cm}{\textbf{Self-supervised Training}} & Rot Prediction\cite{sohn2020learning} &91.3 &84.1&\textbf{95.8} & 86.4\\
& Contrastive\cite{sohn2020learning} &89.0 &82.4&93.6& 87.7\\
& Contrastive+DA\cite{sohn2020learning} &92.5 &86.5&94.8& 89.6\\
&Geometric Trans\cite{golan2018deep} &86.0 &78.7&93.5& 88.8\\
&InvAE\cite{fei2020attribute} &86.6 &78.8&93.9& -\\
& Rot + Trans 
\cite{hendrycks2019using}             & 90.1  & -  & -    & -       \\
                                          & GOAD \cite{bergman2020classification}                     & 88.2  & -  & 94.1     & -       \\
                                          & CSI \cite{tack2020csi}                     & \textbf{94.3}  & 89.6  & -    & -         \\
                                          & SSD \cite{sehwag2021ssd}                       & 90.0  &   -   &    -      & -         \\
\midrule
\multirow{2}{2cm}{\textbf{ImageNet Pre-trained}}              & Supervised    & 86.2  & 87.1  & 90.2     & 89.4           \\
                                          & \textbf{Self-supervised (ours)} & 93.8  & \textbf{92.6} & 94.4     & \textbf{94.8}      \\
\bottomrule
\end{tabular}

\end{table}

The one-class anomaly detection tasks on CIFAR-10 and CIFAR-100 are more commonly studied in previous work, and we present more results on these two tasks, including the per-class AUROCs and confusion matrices in Appendix \ref{additional one-class cifar}.

\subsection{Ablation Study}\label{ablation sec}
In this section, We perform an ablation study on the components of our method introduced in Section \ref{components}. Throughout this section, we conduct experiments on the CIFAR-100 one-class anomaly detection task. We report the corresponding ablation study on the CIFAR-10 OOD detection task in Appendix \ref{additional ablation}, where we found the results to be consistent.

\textbf{Representation Learning Algorithm. }The core component of our method is the feature extractor trained on ImageNet with self-supervised loss. In Table \ref{ablation algo}, we compare several state-of-the-art self-supervised representation learning algorithms while keeping other factors fixed. We observe that all algorithms can be effectively applied to the anomaly detection task. In particular, the algorithms based on the contrastive loss in \eqref{simclr_loss} (SimCLR and MoCo) obtain better performances than algorithms with alternative losses, although the latter may obtain better classification accuracy on ImageNet (under the linear evaluation protocol). Note that previous two-stage outlier detection algorithms, such as \cite{tack2020csi,sehwag2021ssd, sohn2020learning} also train representations with losses similar to \eqref{simclr_loss}. Therefore, probably the contrastive loss introduces a good inductive bias for outlier detection, as it explicitly pushes dissimilar images away. Given its popularity in previous work on outlier detection and the its strong empirical performance in our study, SimCLRv2 is chosen as the pre-train algorithm for all experiments. 

\begin{table}[t]
\begin{minipage}{.49\linewidth}
    \centering
\caption{Ablation study on representation learning algorithms for the pre-trained feature extractor with ResNet-50 structure. Top1 Acc is the top 1 accuracy for linear evaluation on ImageNet.}
\label{ablation algo}
   \begin{tabular}{lll}
   \toprule
      Algorithm & AUROC & Top1 Acc\\
      \midrule
      SimCLRv2 \cite{chen2020big} & \textbf{92.6} & 74.6\\
      MoCov2 \cite{chen2020improved} &92.1 & 71.1\\
      SwAV \cite{caron2020unsupervised} & 91.2& \textbf{75.3} \\
      BYOL \cite{grill2020koray} & 90.7 & 74.3\\
      Barlow Twins \cite{zbontar2021barlow} &89.5& 73.2 \\
      \bottomrule
 \end{tabular}
\end{minipage}\hfill
\begin{minipage}{.5\linewidth}
    \centering
\caption{Ablation study on the network structure for the feature extractor trained with SimCLRv2. Results of linear evaluation accuracy are reported in \cite{chen2020big}.}
\label{ablation net struct}
   \begin{tabular}{lll}
   \toprule
      Network & AUROC & Top1 Acc \\
      \midrule
      ResNet-50 & 92.6 &74.6\\
      ResNet-50 2$\times$ & 87.2& \textbf{77.7}\\
      ResNet-101 & 93.3& 76.3 \\
      ResNet-152 & \textbf{94.0} & 77.2\\
\bottomrule
\end{tabular}
\end{minipage}\hfill
\end{table}

\textbf{Network Structure. }Recent advances in self-supervised representation learning mainly use ResNet-50 as the backbone structure, while wider and deeper variants also exist. We fix the SimCLRv2 learning algorithm, and explore the effect of feature extractor's capacity on outlier detection in Table \ref{ablation net struct}. We observe that deeper networks lead to slightly improved performance. We also note that there is a performance drop on wider ResNet-50, and the reason is that the wider network doubles the dimension of the feature space (4096 vs. 2048), making the feature space outlier detection much harder (note that for each class, we only have 2500 inlier samples). Considering the computational cost in test time, we choose to use the basic ResNet-50 structure throughout the paper.
 
\textbf{Feature Space Outlier Detector. }
We compare different feature space outlier detectors in Table \ref{ablation feature detector}, including non-parametric (OC-SVM with RBF kernel and KDE with Gaussian kernel) and parametric (Mahalanobis distance) detectors. For non-parametric detectors, we do a grid search on the hyper-parameters (such as the kernel coefficient for OC-SVM and the bandwidth for KDE) and report the best results. For Mahalanobis distance described in equation \eqref{maha}, we try either a single component or 4 components obtained from K-means clustering. We do not consider more components because of the small in-lier data size (while in Appendix \ref{additional ablation}, we try more components as the in-lier size is larger). We also compare using a tied covariance matrix vs. component conditional covariance matrices. From Table \ref{ablation feature detector}, we observe that parametric detectors outperforms non-parametric detectors. This suggests that the representations extracted by ImageNet pre-trained network have a compact structure, which can be well approximated by simple parametric models. Interestingly, we find that for Mahalanobis distance, a single tied covariance matrix outperforms a separate covariance matrix for each cluster. Considering its good performances and computational efficiency, we use Mahalanobis distance with a tied covariance matrix as the feature outlier detector in the main experiments. 

\textbf{Input Pre-processing. }We use a single feature extractor for all outlier detection tasks, where images can vary significantly in sizes. We need to pre-process the data to fit the input size of feature extractors trained on ImageNet. In particular, for small image datasets such as CIFAR-10 and CIFAR-100 with resolution $32 \times 32$, we need to up-sample the images by a factor of $8$ (followed by a center crop to size $224 \times 224$). We study the effect of different up-sampling methods in Table \ref{ablation preprocess}, where we observe that more sophisticated up-samplers such as bi-cubic interpolation and Lanczos resampling lead to significantly better results than simple nearest neighbor up-sampling. We use bi-cubic interpolation throughout the main results for a balance of simplicity and effectiveness. We believe that using learning based up-sampler \cite{wang2020deep} will lead to improved performances, however, since our goal is to avoid any training on inlier data, we do not explore that direction. 

Additionally, we study the effectiveness of the input perturbation in equation \eqref{perturbation}. We see in Table \ref{ablation preprocess} that the perturbation does improve the outlier detection result, suggesting that such perturbation can have stronger effect on separating the inlier and outlier samples. However, due to the increased computational burden during the test time, we do not perturb the data when reporting the main results. 

\begin{table}[t]
\begin{minipage}{.49\linewidth}
    \centering
\caption{Ablation study on the feature space outlier detector. Non-parametric and parametric detectors are separated by the bar. \textbf{MD} stands for Mahalanobis distance, and \textbf{tied} means using a tied covariance matrix for all components.}
\label{ablation feature detector}
   \begin{tabular}{lll}
   \toprule
      Decision score &  Component & AUROC \\
      \midrule
      OC-SVM & - & 90.3\\
      KDE &- & 88.7\\
      \midrule
      MD & 1& 91.8\\
      MD & 4 & 86.7\\
      MD (tied) &4 & \textbf{92.6}\\
      \bottomrule
 \end{tabular}
\end{minipage}\hfill
\begin{minipage}{.5\linewidth}
    \centering
\caption{Ablation study on the input pre-processing. Perturbation corresponds to optionally apply equation \eqref{perturbation} on test inputs with $\epsilon = 0.01$.}
\label{ablation preprocess}
   \begin{tabular}{lll}
   \toprule
      Up-sampling & Perturbation & AUROC \\
      \midrule
      Nearest Neighbor & No &80.7\\
      Bilinear  & No& 90.4\\
      Cubic & No& 92.6 \\
      Cubic & Yes& \textbf{93.7} \\
      Lanczos & No & 92.8\\
\bottomrule
\end{tabular}
\end{minipage}\hfill
\end{table}
\section{Conclusion and Discussion}\label{conclusion}
In this paper, we study the effectiveness of using representations pre-trained on ImageNet to detect outliers from various sources. Through extensive experiments, we show that by leveraging a single publicly available pre-trained feature extractor with self-supervised loss on ImageNet, we can achieve competitive performance on various outlier detection tasks, and such effectiveness is consistent across choices on representation learning algorithms and detection scores in the feature space. While previous work highlighted the importance of learning representations from in-domain data, our study suggests that the actual benefits of domain-specific training may be marginal. Our method has important practical implications, as it is easy to use and requires no training of the representation space. In addition, our results can serve as an important baseline for future studies on outlier detection.

Given the on-going rapid progress in self-supervised representation learning, we are confident that the performance of our method can further be improved with advanced algorithm and larger pre-train datasets \cite{mahajan2018exploring,dosovitskiy2021an}. Our method has one obvious limitation: it is not trivial to extend it to domains other than images, as the method requires pre-trained representations on a large reference dataset. However, we believe that such an issue will shortly be resolved, as self-supervised representation learning becomes an increasingly popular topic in different domains \cite{dave2021tclr, saeed2020contrastive,you2020graph,fang2020cert}.

\newpage
\bibliographystyle{plainnat}
\bibliography{reference}

%%%%%%%%%%%%%%%%%%%%%%%%%%%%%%%%%%%%%%%%%%%%%%%%%%%%%%%%%%%%
%%%%%%%%%%%%%%%%%%%%%%%%%%%%%%%%%%%%%%%%%%%%%%%%%%%%%%%%%%%%
\newpage
\appendix
\section{Experimental Details}\label{exp detail}
In this section, we introduce the detailed settings of our main experiments. For all results reported except some results of ablation study, we use the pre-trained ResNet-50 by SimCLRv2 on ImageNet dataset. Details regarding obtaining the pre-trained weight will be discussed in Appendix \ref{download}. We discard the projection head, and use the last layer of the ResNet-50 after average pooling as our feature. The feature is of dimension 2048. Regardless the original size of the images, we first resize the image to $256 \times 256$ and then center-crop to $224 \times 224$. All images are normalized to have pixel value in $[0,1]$ \footnote{This is because the SimCLRv2 pre-train model is implemented and trained with TensorFlow. For ablation study where we use pre-trained models in PyTorch, we follow their own data normalization parameters}.

For each task, we collect the features of inlier images by the feature extractor $f$. Then we use K-means to partition the training features to $M$ clusters and obtain the cluster assignment for each image. Then we compute the cluster conditioned mean for each cluster $m = 1,\dots,M$:
\begin{align*}
    \widehat{\mu}_{m}=\frac{1}{N_{m}} \sum_{i: y_{i}=m} f\left(\mathbf{x}_{i}\right),
\end{align*}
where $N_{m}$ is the number of training samples assigned to cluster $m$ by the K-means algorithms. We also compute the tied covariance matrix 
\begin{align*}
    \widehat{\mathbf{\Sigma}}=\frac{1}{N} \sum_{m} \sum_{i: y_{i}=m}\left(f\left(\mathbf{x}_{i}\right)-\widehat{\mu}_{m}\right)\left(f\left(\mathbf{x}_{i}\right)-\widehat{\mu}_{m}\right)^{\top}.
\end{align*}
The decision score is computed by equation (\ref{maha}) with the tied covariance matrix. 

\textbf{Number of components: }We decide the number of components for each task for a balance between computational complexity and effectiveness. For CIFAR-10 OOD detection task, we set the number of components to be 8; for CIFAR-10, CIFAR-100 and Fashion MNIST one-class anomaly detection task, we set the number of components to be 4. For OOD detection tasks on high-resolution images (results in Table \ref{large image ood}) and the Cat-vs-Dog task, we set the number of components to be 10. 
\subsection{Dataset}
For the datasets we used, when the dataset comes with a train/test split, we use training data for collecting the inlier features, and use testing data use CIFAR-10 and CIFAR-100 (superclass). CIFAR-10 and CIFAR-100 consist of 50,000 training and 10,000
test images with 10 and 20 (super-class) image classes, respectively. 

For OOD detection task on small images, our in-lier dataset is CIFAR-10. The OOD samples are as follows: SVHN consists of 26,032
test images with 10 digits, resized LSUN consists of 10,000 test images of 10 different scenes,
resized ImageNet consists of 10,000 test images with 200 images classes from a subset of full
ImageNet dataset, and LSUN (FIX), ImageNet (FIX) consists of 10,000 test images. 

For datasets we used for OOD detection task on larger images as presented in Table \ref{large image ood}, we randomly choose 3000 images from each dataset as test data, except that the ImageNet-30 comes with a test set of 3000 images.

\section{OOD and Anomaly Detection Benchmark on ImageNet-30}\label{imagenet-30 benchmark}
In Table \ref{imagenet-30 table}, we compare the OOD detection performance of our method and methods based on training self-supervised representations on in-distribution data, when the in-distribution data is ImageNet-30. We observe that our method significantly outperform all other methods. In particular, the feature extractor with self-supervised pre-training significantly outperforms that with supervised pre-training. Note that ImageNet-30 is a subset of ImageNet, and we use the self-supervised feature extractor pre-trained on the whole ImageNet, so it is expected that our method will perform well in this task.

\begin{table}

  \caption{AUROC ($\%$) of various unsupervised OOD detection methods with ImageNet-30 as in-distribution.}
  \label{imagenet-30 table}
  \centering
 \begin{tabular}{lllllllll}
 \toprule
                            &             & CUB                          & Dogs                         & Flowers                                               & Pets                         & Places                       & Caltech                   \\ \midrule
                       \multirow{3}{3cm}{\textbf{Self-supervised Training}} &         Rot + Trans \cite{hendrycks2019using} & 74.5 & 77.8 & 86.3& 70.0 & 53.1 & 70.0 \\
                      &  GOAD \cite{bergman2020classification} & 71.5 & 74.3 & 82.8 & 65.5 & 51.0 & 67.4 \\
                    & CSI \cite{tack2020csi} & 90.5 & 97.1 & 94.7 & 85.2 & 78.3 & 87.1 \\
                    \midrule
      \multirow{2}{2cm}{\textbf{ImageNet Pre-trained}}  &              Supervised & 74.5 & 95.4 & 85.1 & 92.5 & 75.4 &  80.9 \\
                   & Self-supervised & \textbf{99.8} & \textbf{99.5} & \textbf{98.1} & \textbf{99.7} &  \textbf{80.0} & \textbf{89.2}\\
                             \bottomrule
\end{tabular}
\end{table}

Similarly, in Table \ref{imagenet-30 anomaly}, we tackle the one-class anomaly detection task on ImageNet-30 with our method, and we observe that ImageNet pre-trained feature extractors, with either supervised or self-supervised algorithms, obtain significantly better results than previous baselines. This is also expected, as representations extracted by supervised/self-supervised representation learning algorithms achieve high accuracy on linear classification on ImageNet.

\begin{table}
  \caption{Mean AUROC ($\%$) for one-class classification AUCs averaged over outlier classes on ImagNet-30.}
  \label{imagenet-30 anomaly}
  \centering
 \begin{tabular}{lllllll}
 \toprule
                                          &                           & ImageNet-30 \\
\midrule
 \multirow{3}{3cm}{\textbf{Self-supervised Training}} & Rot + Trans 
\cite{hendrycks2019using}             & 77.9  \\
& Rot+Trans+Attn+Resize
\cite{hendrycks2019using}             & 84.8  \\                                         
                                          & CSI \cite{tack2020csi}                     & 91.6  \\
                                        
\midrule
\multirow{2}{2cm}{\textbf{ImageNet Pre-trained}}              & Supervised    & \textbf{99.8}       \\
                                          & \textbf{Self-supervised (ours)} & 99.4     \\
\bottomrule
\end{tabular}

\end{table}

\section{Additional One-class Anomaly Detection Results on CIFAR-10 and CIFAR-100}\label{additional one-class cifar}
The one-class anomaly detection tasks on CIFAR-10 and CIFAR-100 are studied extensively in previous work. Due to the space limit, in this section we compare our method with previous method in details. In Table \ref{table cifar10 perclass} we compare our method with other methods (based on generative models, one-class classifiers and representation learning) on one-class anomaly detection task on CIFAR-10 for each of the 10 classes. Our method is competitive with CSI, the current SOTA. Each of our method and CIS obtains the best result on 5 out of 10 classes, while on average our method is only slightly worse (93.8 vs. 94.3).

In Table \ref{table oc-confusion}, we present the confusion matrix of AUROC of our method on one-class anomaly detection task on CIFAR-10, where bold denotes the hard pairs (AUROC less than 80$\%$). The results align with the human intuition that ‘car’ is confused to ‘truck’, ‘cat’ is confused to ‘dog’, and ‘deer’ is confused to ‘horse’.

Table \ref{table oc cifar100} compares our method with various methods on one-class anomaly detection task on CIFAR-100 (super-class), for each of the 20 super-classes. Our method outperforms the prior methods for most of the classes and obtains the best average result.

\begin{table}[!htb]
\centering
\footnotesize
\caption{Comparison of our method with other detectors for one class anomaly detection task on CIFAR-10. Results of other methods are presented in ~\citep{sehwag2021ssd}.}
\label{table cifar10 perclass}
\renewcommand{\arraystretch}{1.1}
\resizebox{0.9\linewidth}{!}{
\begin{tabular}{cccccccccccc} 
\toprule
 & Airplane & Automobile & Bird & Cat & Deer & Dog & Frog & Horse & Ship & Truck & Average \\ \midrule
Randomly Initialized network & 77.4 & 44.1 & 62.4 & 44.1 & 62.1 & 49.6 & 59.8 & 48.0 & 73.8 & 53.7 & 57.5 \\
VAE & 70.0 & 38.6 & 67.9 & 53.5 & 74.8 & 52.3 & 68.7 & 49.3 & 69.6 & 38.6 & 58.3 \\
PixelCNN & 53.1 & 99.5 & 47.6 & 51.7 & 73.9 & 54.2 & 59.2 & 78.9 & 34.0 & 66.2 & 61.8 \\
OCSVM~\citep{scholkopf1999support} & 63.0 & 44.0 & 64.9 & 48.7 & 73.5 & 50.0 & 72.5 & 53.3 & 64.9 & 50.8 & 58.5 \\
AnoGAN~\citep{schlegl2017unsupervised} & 67.1 & 54.7 & 52.9 & 54.5 & 65.1 & 60.3 & 58.5 & 62.5 & 75.8 & 66.5 & 61.8 \\

DSVDD~\citep{ruff2019deep} & 61.7 & 65.9 & 50.8 & 59.1 & 60.9 & 65.7 & 67.7 & 67.3 & 75.9 & 73.1 & 64.8 \\
OCGAN~\citep{perera2019ocgan} & 75.7 & 53.1 & 64.0 & 62.0 & 72.3 & 62.0 & 72.3 & 57.5 & 82.0 & 55.4 & 65.6 \\
RCAE~\citep{chalapathy2018anomaly} & 72.0 & 63.1 & 71.7 & 60.6 & 72.8 & 64.0 & 64.9 & 63.6 & 74.7 & 74.5 & 68.2 \\
DROCC~\citep{goyal2020drocc} & 81.7 & 76.7 & 66.7 & 67.1 & 73.6 & 74.4 & 74.4 & 71.4 & 80.0 & 76.2 & 74.2 \\
Deep-SAD~\citep{ruff2019deep} & -- & -- & -- & -- & -- & -- & -- & -- & -- & -- & 77.9 \\
E3Outlier~\citep{wang2019effective} & 79.4 & 95.3 & 75.4 & 73.9 & 84.1 & 87.9 & 85.0 & 93.4 & 92.3 & 89.7 & 85.6 \\
Geom Trans~\citep{golan2018deep} & 74.7 & 95.7 & 78.1 & 72.4 & 87.8 & 87.8 & 83.4 & 95.5 & 93.3 & 91.3 & 86.0 \\
InvAE~\citep{fei2020attribute} & 78.5 & 89.8 & 86.1 & 77.4 & 90.5 & 84.5 & 89.2 & 92.9 & 92.0 & 85.5 & 86.6 \\
GOAD~\citep{bergman2020classification} & 77.2 & 96.7 & 83.3 & 77.7 & 87.8 & 87.8 & 90.0 & 96.1 & 93.8 & 92.0 & 88.2 \\
CSI~\citep{tack2020csi} & 89.9 & \textbf{99.9} & \textbf{93.1} & 86.4 & \textbf{93.9} & 93.2 & 95.1 & \textbf{98.7} & \textbf{97.9} & 95.5 & \textbf{94.3} \\ 
SSD ~\citep{sehwag2021ssd} & 82.7 & 98.5 & 84.2 & 84.5 & 84.8 & 90.9 & 91.7 & 95.2 & 92.9 & 94.4 & 90.0 \\
\midrule
Supervised pre-train & 84.5 & 96.1 & 77.3 & 78.9 & 84.8 & 82.3 & 90.7 &88.6 & 85.3 & 94.5 & 86.2\\
Ours & \textbf{94.8} & 96.4 & 88.3 & \textbf{87.6} & 92.7 & \textbf{94.2} & \textbf{96.4} & 94.3 & 96.1 & \textbf{97.0} & 93.8\\
\bottomrule
\end{tabular}}
\end{table}

\begin{table}[h]
\centering\small
\caption{
Confusion matrix of AUROC (\%) values of our method on one-class CIFAR-10. The row and column indicates the in-distribution and OOD class, respectively, and the final column indicates the mean value. Bold denotes the values under 80\%, which implies the hard pair.
}\label{table oc-confusion}
\begin{tabular}{lcccccccccc|c}
\toprule
& Plane & Car & Bird & Cat & Deer & Dog & Frog & Horse & Ship & Truck & Mean \\
\midrule
Plane & - & 94.5 & 95.7	 & 98.9	& 97.4	&99.3	&99.1	& 96.8	&81.1	&91.4&95.0\\
Car   &97.2 &-&	99.8 & 99.7 & 99.6 & 99.8 &	99.8	& 99.6 & 96.0	& \textbf{76.5}&96.5\\
Bird  & 87.8 & 99.5		& -& 90.5	&\textbf{63.5} &	90.5	& 87.0	& \textbf{79.7}&	98.0 & 99.2&88.4\\
Cat   & 95.4&98.5&91.1 & - & 81.7&	\textbf{58.4} & 82.9	&83.1	& 98.1	& 97.9 & 87.4\\
Deer  & 96.6	& 99.7 &	90.2 &	93.8 &-&	93.5	& 92.3 &	\textbf{69.2}	&98.9	& 99.3 &92.6\\
Dog   & 99.3 & 99.7	&97.5  &81.9	& 90.4	&-	&96.1&	85.3	&99.7	& 99.6&94.3\\
Frog  & 98.7	&99.7	&94.5	&93.3	&89.0	&95.8	&-&	97.6	&99.7	&99.8 &96.4
 \\
Horse & 96.9 &99.2 &96.4	&95.1	&\textbf{75.7}	& 91.9 &98.3 &-&	99.2	&98.2 &94.5
 \\
Ship  & 86.2 &92.5	&99.2	&99.4	&98.9	&99.6	&99.7	&99.0&-		&90.0&96.0
 \\
Truck & 96.4 & \textbf{79.5} & 99.9 & 99.8 & 99.9 & 99.9 & 99.9 & 99.3 & 96.5 & - & 96.9 \\
\bottomrule
\end{tabular}
\end{table}

\begin{table*}[h]
\vspace{-0.05in}
\centering\small
\caption{
AUROC (\%) values of various OOD detection methods trained on one-class CIFAR-100 (super-class). Each row indicates the results of the selected super-class, and the final row indicates the mean value. The values of other methods are from the reference.
}\label{table oc cifar100}
\resizebox{\textwidth}{!}{
\begin{tabular}{ccccccccccc}
\toprule
& OC-SVM & DAGMM & DSEBM & ADGAN & Geom Trans  & Rot+Trans  & GOAD  & CSI & Ours \\
\midrule
0 & 68.4 & 43.4 & 64.0 & 63.1 & 74.7 & 79.6 & 73.9 & 86.3& \textbf{91.6}  \\
1 & 63.6 & 49.5 & 47.9 & 64.9 & 68.5 &  73.3 & 69.2 & 84.8 &\textbf{88.3} \\
2 & 52.0 & 66.1 & 53.7 & 41.3 & 74.0 &  71.3 & 67.6 & 88.9&\textbf{95.0}\\
3 & 64.7 & 52.6 & 48.4 & 50.0 & 81.0 &  73.9 & 71.8 & 85.7& \textbf{96.7} \\
4 & 58.2 & 56.9 & 59.7 & 40.6 & 78.4 &  79.7 & 72.7 & 93.7& \textbf{94.8}\\
5 & 54.9 & 52.4 & 46.6 & 42.8 & 59.1 & 72.6 & 67.0 & 81.9&\textbf{95.1}\\
6 & 57.2 & 55.0 & 51.7 & 51.1 & 81.8 &  85.1 & 80.0 & 91.8 &\textbf{96.5} &\\
7 & 62.9 & 52.8 & 54.8 & 55.4 & 65.0 &  66.8 & 59.1 & 83.9 &\textbf{90.0} \\
8 & 65.6 & 53.2 & 66.7 & 59.2 & 85.5 &  86.0 & 79.5 & 91.6 &\textbf{93.0} \\
9 & 74.1 & 42.5 & 71.2 & 62.7 & 90.6 &  87.3 & 83.7 & \textbf{95.0}&92.5\\
10 & 84.1 & 52.7 & 78.3 & 79.8 & 87.6 &  88.6 & 84.0 & 94.0&\textbf{95.2}\\
11 & 58.0 & 46.4 & 62.7 & 53.7 & 83.9 &  77.1 & 68.7 & 90.1&\textbf{91.6}\\
12 & 68.5 & 42.7 & 66.8 & 58.9 & 83.2 &  84.6 & 75.1 & \textbf{90.3} & \textbf{90.3}\\
13 & 64.6 & 45.4 & 52.6 & 57.4 & 58.0 &  62.1 & 56.6 & 81.5&\textbf{85.5}\\
14 & 51.2 & 57.2 & 44.0 & 39.4 & 92.1 &  88.0 & 83.8 & 94.4&\textbf{96.7}\\
15 & 62.8 & 48.8 & 56.8 & 55.6 & 68.3 &  71.9 & 66.9 & 85.6&\textbf{86.5}\\
16 & 66.6 & 54.4 & 63.1 & 63.3 & 73.5 &  75.6 & 67.5 & 83.0&\textbf{88.6}\\
17 & 73.7 & 36.4 & 73.0 & 66.7 & 93.8 &  93.5 & 91.6 & \textbf{97.5} & 95.6\\
18 & 52.8 & 52.4 & 57.7 & 44.3 & 90.7 &  91.5 & 88.0 & \textbf{95.9} & 95.3\\
19 & 58.4 & 50.3 & 55.5 & 53.0 & 85.0 &  88.1 & 82.6 & 95.2 & \textbf{93.1}\\
\midrule
Mean & 63.1 & 50.6 & 58.8 & 55.2 & 78.7 & 79.8 & 74.5 & 89.6 & \textbf{92.6}\\
\bottomrule
\end{tabular}}
\end{table*}

\section{Additional Ablation Study}\label{additional ablation}
In this section, we provide results of ablation studies corresponding to Section \ref{ablation sec} on the CIFAR-10 OOD detection experiments. In Table \ref{cifar ood ablation 1}, \ref{cifar ood ablation 2}, \ref{cifar ood ablation 3} and \ref{cifar ood ablation 4}, we present the results of ablation study on representation learning algorithm, network structure, feature space outlier detector and data pre-processing respectively. The findings are overally consistent to those in Section \ref{ablation sec}.

\begin{table}
  \caption{Ablation study on representation learning algorithms for the pre-trained feature extractor with ResNet-50 structure for the CIFAR-10 OOD detection task.}
  \label{cifar ood ablation 1}
  \centering \small
  \resizebox{\textwidth}{!}{
 \begin{tabular}{llllllll}
 \toprule
                                                                     & SVHN  & LSUN  & ImageNet & LSUN (F) & ImageNet (F) &CIFAR-100 \\
\midrule
  SimCLRv2\cite{chen2020big}                     & \textbf{98.3}  & \textbf{98.5}  & \textbf{98.6} & \textbf{92.9}& \textbf{91.8}& \textbf{81.7}     \\
                                           MoCov2 \cite{chen2020improved}                    & 96.3  &    89.4  &    89.7  &86.7&86.0  & 76.2         \\
                                           SwAV \cite{caron2020unsupervised}                    & 95.5  &  90.8    &  91.9     &81.7&82.5   & 74.1         \\
                                          BYOL\cite{grill2020koray}         &90.5    & 83.3   &85.1&80.9  & 81.3    & 74.8        \\
                                          Barlow Twins\cite{zbontar2021barlow}    & 94.7  &  90.5    &  90.8       &87.4&87.1&     77.9            \\
\bottomrule
\end{tabular}}
\end{table}

\begin{table}
  \caption{Ablation study on the network structure for feature extractor trained with SimCLRv2 for CIFAR-10 OOD detection task.}
  \label{cifar ood ablation 2}
  \centering \small
  \resizebox{\textwidth}{!}{
 \begin{tabular}{llllllll}
 \toprule
                                                                     & SVHN  & LSUN  & ImageNet & LSUN (F) & ImageNet (F) &CIFAR-100 \\
\midrule
  ResNet-50  & 98.3  &98.5  & 98.6 &92.9& 91.8& 81.7      \\
                                           ResNet-50 2$\times$               & 83.0  & 90.2  &    91.4   &81.5&82.3  & 71.7         \\
                                          ResNet-101                 & 97.9  &  99.2   &  \textbf{99.1}     &90.3&91.1   & 82.0         \\
                                          ResNet-152 & \textbf{98.5}   &\textbf{99.3}&99.0  & \textbf{92.6} &  \textbf{92.1}&\textbf{82.3}            \\
\bottomrule
\end{tabular}}
\end{table}

\begin{table}
  \caption{Ablation study on the feature space outlier detector for the CIFAR-10 OOD detection task.}
  \label{cifar ood ablation 3}
  \centering \small
  \resizebox{\textwidth}{!}{
 \begin{tabular}{llllllll}
 \toprule
                                                                     & SVHN  & LSUN  & ImageNet & LSUN (F) & ImageNet (F) &CIFAR-100 \\
\midrule
  OC-SVM                & 93.7       & 96.3    &96.7& 82.1& 84.5&76.9         \\
                                           KDE               & 93.4  &    95.5   & 96.0  &80.8&83.7  & 76.5         \\
                        \midrule                 MD-1 component                    & 95.0  &  98.0   &  97.6     &91.4&89.2   & 80.1         \\
                                          MD-4 component& 95.1   &98.4  & \textbf{98.8} &92.6&  90.3 & 78.9        \\
                     MD-4 component (tied)&96.4  &\textbf{99.0}&98.7  & 91.5 & 89.7 & 81.0            \\
                     MD-8 component (tied)  & \textbf{98.3}  &98.5  & 98.6 &\textbf{92.9}& \textbf{91.8}& \textbf{81.7}          \\
\bottomrule
\end{tabular}}
\end{table}

\begin{table}
  \caption{Ablation study on the input pre-processing for CIFAR-10 OOD detection task. Perturbation corresponds to optionally apply equation \eqref{perturbation} on test inputs with $\epsilon = 0.01$.}
  \label{cifar ood ablation 4}
  \centering \small
  \resizebox{\textwidth}{!}{
 \begin{tabular}{llllllll}
 \toprule
                                                                     & SVHN  & LSUN  & ImageNet & LSUN (F) & ImageNet (F) &CIFAR-100 \\
\midrule
  Nearest Neighbor                  & 85.7   &85.8     & 85.4    &61.6& 68.7& 65.2         \\
                                          Bilinear               &   92.7  &96.6  & 95.7 &88.8& 87.3& 78.7         \\
                                          Cubic                  & 98.3  &98.5  & 98.6 &\textbf{92.9}& 91.8& 81.7          \\
                                          Cubic + perturb & \textbf{99.3} & \textbf{99.1}   &\textbf{99.0}&92.5  & \textbf{92.0}& \textbf{82.1}         \\
                    Lanczos &98.2&99.5&99.5&91.5 & 89.9& 81.0\\
\bottomrule
\end{tabular}}
\end{table}

\section{Instructions for downloading pre-trained feature extractors}\label{download}
In this section, we provide instructions for downloading the publicly available pre-trained feature extractors mentioned in our paper. We have read the license carefully before using the publicly released models.

\textbf{Supervised pre-trained ResNet-50: }We use the official pre-trained ResNet-50 model in torchvision. See \url{https://pytorch.org/vision/stable/models.html} for details. The URL to the pre-trained weight is \url{https://download.pytorch.org/models/resnet50-19c8e357.pth}.

\textbf{SimCLRv2: }See \url{https://github.com/google-research/simclr} for detailed implementations and license. The pre-trained models are in TensorFlow format, and following the provided instruction, we use the codes at \url{https://github.com/Separius/SimCLRv2-Pytorch} to convert the weights into PyTorch format. 

\textbf{MoCov2: }See \url{https://github.com/facebookresearch/moco} for implementations and license. The particular model weight we used can be downloaded at \url{https://dl.fbaipublicfiles.com/moco/moco_checkpoints/moco_v2_800ep/moco_v2_800ep_pretrain.pth.tar}

\textbf{SwAV: }See \url{https://github.com/facebookresearch/swav} for implementations and license. The particular model weight we used can be downloaded at \url{https://dl.fbaipublicfiles.com/deepcluster/swav_800ep_pretrain.pth.tar}

\textbf{BYOL: }We use the PyTorch implementation of BYOL at \url{https://github.com/yaox12/BYOL-PyTorch}. Pre-trained weight can be found at \url{https://drive.google.com/file/d/1TLZHDbV-qQlLjkR8P0LZaxzwEE6O_7g1/view?usp=sharing}

\textbf{Barlow Twins: }See \url{https://github.com/facebookresearch/barlowtwins} for implementations and license. The particular model weight we used can be downloaded at \url{https://dl.fbaipublicfiles.com/barlowtwins/epochs1000_bs2048_lr0.2_lambd0.0051_proj_8192_8192_8192_scale0.024/resnet50.pth}
\end{document}